\documentclass[conference]{IEEEtran}

\usepackage{PRIMEarxiv}

\usepackage[utf8]{inputenc} 
\usepackage[T1]{fontenc}   

\usepackage{url}           
\usepackage{booktabs}       
\usepackage{amsfonts}       
\usepackage{nicefrac}      
\usepackage{microtype}      
\usepackage{lipsum}

\usepackage{graphicx}      
\graphicspath{{media/}}     

\usepackage{xcolor}  
\usepackage{hyperref}
\usepackage{amsmath}
\usepackage{amsthm}
\usepackage{amssymb}
\usepackage{mathtools}
\usepackage{stmaryrd}
\usepackage[ruled,lined,linesnumbered,boxed]{algorithm2e}
\usepackage{bbm}
\usepackage{multirow}

\newtheorem{proposition}{Proposition}

\DeclareMathOperator*{\argmax}{arg\,max}

\begin{document}

\title{Enhanced Estimation Techniques for Certified Radii in Randomized Smoothing}
\author{\IEEEauthorblockN{Zixuan Liang}
\IEEEauthorblockA{\textit{zliang1@my.harrisburgu.edu}}
}

    \maketitle

   \begin{abstract}
       This paper presents novel methods for estimating certified radii in randomized smoothing, a technique crucial for certifying the robustness of neural networks against adversarial perturbations. Our proposed techniques significantly improve the accuracy of certified test-set accuracy by providing tighter bounds on the certified radii. We introduce advanced algorithms for both discrete and continuous domains, demonstrating their effectiveness on CIFAR-10 and ImageNet datasets. The new methods show considerable improvements over existing approaches, particularly in reducing discrepancies in certified radii estimates. We also explore the impact of various hyperparameters, including sample size, standard deviation, and temperature, on the performance of these methods. Our findings highlight the potential for more efficient certification processes and pave the way for future research on tighter confidence sequences and improved theoretical frameworks. The study concludes with a discussion of potential future directions, including enhanced estimation techniques for discrete domains and further theoretical advancements to bridge the gap between empirical and theoretical performance in randomized smoothing. 
   \end{abstract}


  \section{Introduction}\label{sec:introduction} 
Deep neural networks (DNNs) have transformed fields such as computer vision~\cite{krizhevsky2012imagenet} and natural language processing~\cite{devlin2018bert}, excelling in tasks by learning complex patterns~\cite{lecun2015deep}. Their integration into critical systems, including smartphones and autonomous vehicles, has highlighted significant security concerns~\cite{goodfellow2016deep, biggio2018wild}. Addressing these vulnerabilities is crucial, particularly in adversarial machine learning, which focuses on defending against malicious attacks~\cite{vorobeychik2018adversarial}.  

Adversarial examples, subtly perturbed inputs that mislead DNNs~\cite{szegedy2013intriguing, goodfellow2014explaining}, pose threats across domains like image classification~\cite{carlini2017towards}, speech recognition~\cite{carlini2018audio}, and autonomous systems~\cite{kurakin2016adversarial, eykholt2018robust}. In healthcare, adversarial manipulations can cause misdiagnoses~\cite{finlayson2019adversarial}, while in finance, they could undermine fraud detection and trading algorithms~\cite{gu2018adversarial, li2020adversarial}.  

Defense strategies include adversarial training~\cite{tramer2017ensemble}, input preprocessing~\cite{guo2017countering}, and detection~\cite{metzen2017detecting}. Despite progress, these methods face challenges like increased computational costs and limited generalization~\cite{tsipras2018robustness, xu2017feature}. Certifiable robustness approaches offer formal guarantees, leveraging methods such as MILP~\cite{tjeng2017evaluating}, SMT~\cite{katz2017reluplex}, and randomized smoothing~\cite{cohen2019certified}. While scalable, these techniques often yield conservative bounds~\cite{raghunathan2018semidefinite, wong2018provable}.  

Emerging probabilistic certifications like randomized smoothing balance scalability and rigor, making them promising for state-of-the-art architectures~\cite{cohen2019certified}. As DNNs become ubiquitous, securing them requires a multifaceted approach, blending empirical defenses, formal verification, and innovative designs. The following sections explore randomized smoothing in detail, emphasizing its theoretical underpinnings and practical applications.

In summary, this paper is structured as follows:
\begin{itemize}
    \item Section~\ref{sec:background-&-related-work} introduces the theoretical foundations of randomized smoothing, along with the notations and definitions used throughout.
    \item Section~\ref{sec:problem-formulation} precisely formulates the problem, dividing it into two subproblems: the discrete case and the continuous case.
    \item Section~\ref{sec:discrete} addresses the discrete case, presenting our initial contributions.
    \item Section~\ref{sec:continuous} begins with a novel formulation of confidence intervals in Subsection~\ref{subsec:estimating-by-betting} and extends the discussion to contributions for the continuous case.
    \item Section~\ref{sec:experiments} validates the theoretical results on real-world datasets, including CIFAR-10 and ImageNet.
\end{itemize}

\section{Background \& Related Work}\label{sec:background-&-related-work}

\subsection{Notation}\label{subsec:notation}
Following~\cite{delattre2023lipschitz}, let $\mathcal{X} \subset \mathbb{R}^d$ denote the input space and $\mathcal{Y} = \{1, \ldots, m\}$ represent the set of $m$ class labels.  
For a data point $x \in \mathcal{X}$ with true label $y \in \mathcal{Y}$, a neural network classifier $F_{out}: \mathcal{X} \rightarrow \mathcal{Y}$ predicts as:  
\[
F_{out}(x) = \arg\max_{k \in \mathcal{Y}} s_k \circ f(x),
\]
where $f: \mathcal{X} \rightarrow \mathbb{R}^m$ outputs logits and $s: \mathbb{R}^m \rightarrow \Delta^{m-1}$ is the normalizing layer projecting onto the $(m-1)$-dimensional probability simplex:
\[
\Delta^{m-1} \coloneqq \left\{ p \in \mathbb{R}^m : p_i \geq 0, \sum_{i=1}^m p_i = 1 \right\}.
\]

The composed classifier $F \coloneqq s \circ f$ outputs a probability distribution over $m$ classes, termed the \textbf{soft classifier}, while $F_{out}$ is the \textbf{hard classifier}. Common simplex maps include softmax~\cite{bridle1989training}, hardmax (assigning all mass to the maximum component), and sparsemax~\cite{martins16}, which is the Hilbert projection onto the simplex.  
We refer to the hardmax-based case as the \textbf{discrete case} and cases using continuous projections (e.g., softmax or sparsemax) as the \textbf{continuous case}.

\subsection{Randomized Smoothing}\label{subsec:randomized-smoothing}
Randomized smoothing enhances classifier robustness against small adversarial perturbations by stabilizing decisions under random input noise. This concept, formalized via the robustness radius in Section~\ref{subsec:robustness-radius}, intuitively implies that stable predictions under perturbation yield higher robustness.  

\begin{figure}[h]
    \centering
    \includegraphics[width=0.8\linewidth]{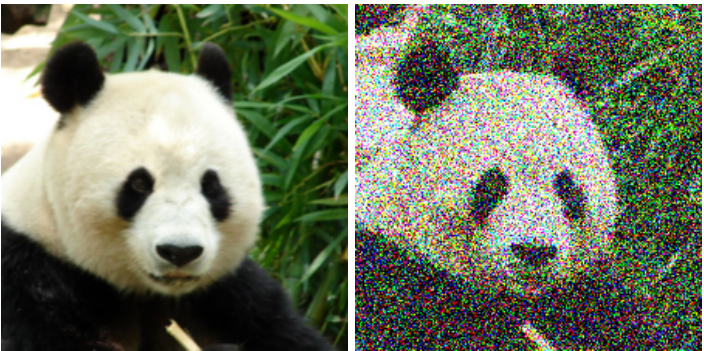}
    \caption{Randomized smoothing: Left, original image $x$ (panda). Right, image with Gaussian noise ($\sigma = 0.5$). The smoothed classifier predicts based on majority voting over noisy samples, mitigating smaller adversarial attacks.}
    \label{fig:pandas}
\end{figure}

Figure~\ref{fig:pandas} from~\cite{cohen2019certified} illustrates this process. The left image depicts a clean input $x$ (a panda image), while the right image shows its perturbed version, generated by applying Gaussian noise with $\sigma = 0.5$. The smoothed classifier aggregates predictions over such noisy samples to improve robustness.

Randomized smoothing enhances robustness by stabilizing predictions under random Gaussian perturbations. The smoothed classifier $\hat{F}: \mathcal{X} \rightarrow \Delta^{m-1}$ implements a majority vote over soft classifier predictions under noise:

\[
    \hat{F}(x) \coloneqq \mathbb{E}_{\epsilon \sim \mathcal{N}(0, \sigma^2 I)} [F(x + \epsilon)],
\]
where $\epsilon \sim \mathcal{N}(0, \sigma^2 I)$ represents isotropic Gaussian noise. The final classifier is:
\[
    \hat{F}_{out}(x) = \arg\max_{k \in \mathcal{Y}} \hat{F}_k(x).
\]
This expectation is equivalent to convolving $F$ with the Gaussian kernel $p_\sigma$, where:
\[
    p_\sigma(z) \coloneqq \frac{1}{(2\pi\sigma^2)^{d/2}} \exp\left(-\frac{\|z\|^2}{2\sigma^2}\right).
\]

\subsection{Robustness Radius}\label{subsec:robustness-radius}
The certified radius $R(F,x)$ quantifies robustness, representing the maximum perturbation $\epsilon$ that keeps the classifier output consistent. Larger $R(F,x)$ implies greater robustness.

\begin{figure}[h]
    \centering
    \includegraphics[width=0.8\linewidth]{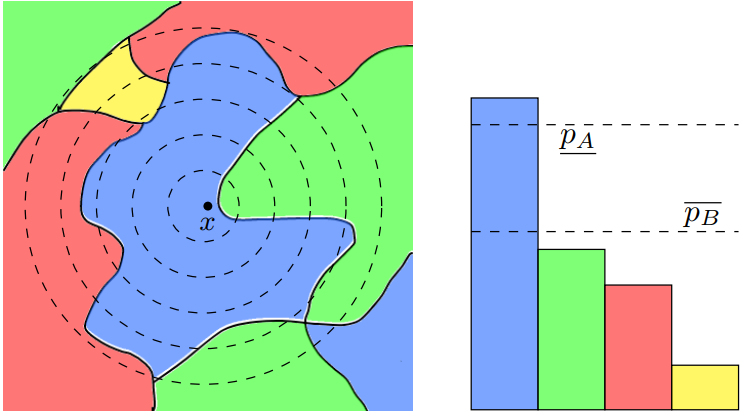}
    \caption{Robustness radius: Left, decision boundaries with perturbation radii around $x$. Right, class probabilities vs. radius. $\underline{p_A}$ and $\overline{p_B}$ denote bounds for top-two class probabilities.}
    \label{fig:radius}
\end{figure}

For randomized smoothing with $\ell_2$-norm, the certified radius is:
\[
    R(F,x) =
    \begin{cases}
        0, & \text{if } F_{out}(x) \neq y, \\[2ex]
        \min \left\{ \epsilon > 0 \,|\, \exists \tau \in B_2(0,\epsilon), F_{out}(x + \tau) \neq y \right\}, & \text{otherwise.}
    \end{cases}
\]

\subsection{Certified Radius and Lipschitz Constant}
Lipschitz continuity relates to robustness: for a Lipschitz constant $L$, 
\[
    \lVert F(x_1)- F(x_2) \rVert \leq L \cdot \lVert  x_1- x_2\rVert.
\]
The prediction margin $M(F,x)$ quantifies confidence:
\[
    M(F,x) \coloneqq F(x)_{y} - \max_{i \neq y}\{F(x)_i\}.
\]
The certified radius is bounded as:
\[
    R(F,x) \geq \frac{M(F,x)}{\sqrt{2}L(F)}.
\]

For a smoothed classifier $\hat{F}$, Lipschitz continuity provides:
\[
    R_1(\hat{F},x) \coloneqq \frac{M(\hat{F},x)}{\sqrt{2}L(f)}.
\]

\subsection{Certified Radius and Noise Magnitude}
Cohen et al.~define a second radius for randomized smoothing:
\[
    R_2(\hat{F},x) = \frac{\sigma}{2} \Big( \Phi^{-1}\big( \hat{F}(x)_{y} \big) - \Phi^{-1} \big( \max_{i \neq y} \{ \hat{F}(x)_i \} \big) \Big),
\]
where $\Phi$ is the Gaussian CDF. Larger $\sigma$ increases robustness but may reduce accuracy on clean inputs, presenting a trade-off.

\section{Problem Formulation}\label{sec:problem-formulation}

\subsection{Monte Carlo Simulation}\label{subsec:monte-carlo-simulation}

Certified radii~\eqref{eq:first-radius} and~\eqref{eq:second-radius} depend on evaluating the smoothed classifier $\hat{F}$, which involves intractable integrals. Monte Carlo methods approximate this by generating Gaussian noise samples, perturbing $x$, and evaluating the base classifier $F$. This approximation is scalable to high-dimensional data and large networks. Using the law of large numbers, the empirical distribution of outputs converges to the true smoothed classifier's distribution as the number of samples $n$ increases.

\subsubsection{Discrete Case}\label{subsubsec:discrete-case-monte-carlo-simulation}
For discrete classifiers, $F$ outputs one-hot vectors (class labels). A vector of counts $X = (X_1, \ldots, X_m)$ is generated by sampling $n$ noise instances $\epsilon_i \sim \mathcal{N}(0, \sigma^2 I)$, perturbing $x$, and evaluating $F$. Algorithm~\ref{alg:vector-of-counts} outlines this. $X$ approximates a multinomial distribution with $n$ trials and probabilities $p = (p_1, \ldots, p_m)$, where $p_k = \mathbb{P}(f(x + \epsilon) = k)$. The final prediction is $\hat{F}_{out}(x) = \argmax_k X_k$. Confidence intervals for $p_k$ (e.g., Clopper-Pearson) ensure robustness by bounding class probabilities.

\begin{algorithm}[h]
    \DontPrintSemicolon
    \KwIn{Base classifier $F$, number of samples $n$, standard deviation $\sigma$, and input $x$}
    \KwOut{Vector of counts $X = (X_1, \ldots, X_m)$}
    Initialize $X = (0, \ldots, 0)$\;
    \For{$i = 1$ \KwTo $n$}{
        Generate noise $\epsilon_i \sim \mathcal{N}(0, \sigma^2 I)$\;
        $\tilde{x}_i = x + \epsilon_i$\;
        $y_i = F(\tilde{x}_i)$\;
        Increment $X_{y_i}$ by 1\;
    }
    \KwRet{$X$}
    \caption{Sampling in the Discrete Case}\label{alg:vector-of-counts}
\end{algorithm}

\subsubsection{Continuous Case}\label{subsubsec:continuous-case-monte-carlo-simulation}
For continuous classifiers \( F: \mathbb{R}^d \rightarrow \Delta^{m-1} \), a matrix \( X \in \mathbb{R}^{n \times m} \) is generated, where each row represents the classifier's output for a noisy sample. Algorithm~\ref{alg:probability-matrix} outlines the process.

The mean confidence for class \( k \) is computed as:
\[
    \hat{p}_k = \frac{1}{n} \sum_{i=1}^n X_i^k
\]
This approximates the smoothed classifier's probabilities. The predicted output of the smoothed classifier is:
\[
    \hat{F}_{out}(x) = \arg\max_k \hat{p}_k
\]
Confidence intervals are incorporated to account for variability in class probabilities, ensuring robust statistical estimates.

\begin{algorithm}[h]
    \DontPrintSemicolon
    \KwIn{Base classifier $F: \mathbb{R}^d \rightarrow \Delta^{m-1}$, number of samples $n$, standard deviation $\sigma$, and input $x$}
    \KwOut{Matrix $X \in \mathbb{R}^{n \times m}$}
    Initialize $X$ as an empty $n \times m$ matrix\;
    \For{$i = 1$ \KwTo $n$}{
        Generate noise $\epsilon_i \sim \mathcal{N}(0, \sigma^2 I)$\;
        $\tilde{x}_i = x + \epsilon_i$\;
        $y_i = F(\tilde{x}_i)$\;
        Set the $i$-th row of $X$ to $y_i$\;
    }
    \KwRet{$X$}
    \caption{Sampling in the Continuous Case}\label{alg:probability-matrix}
\end{algorithm}

\subsection{Monte Carlo Approximation and Confidence Intervals}\label{subsec:monte-carlo-approximation-and-conservative-confidence-intervals}

Monte Carlo introduces randomness into robustness certification, transforming the classifier into a stochastic model. Outputs are random vectors, enabling probabilistic analysis of variability and confidence. To avoid overestimation of robustness, one-sided lower bounds on the certified radius are used. This ensures conservative guarantees even at the cost of slightly smaller radii.

Key trade-offs include:
- **Sample Size ($n$):** Larger $n$ reduces confidence interval width, improving robustness estimates but increases computation.
- **Confidence Level:** Higher levels (e.g., 99\%) yield stronger guarantees but wider intervals and smaller radii.

Both discrete and continuous cases benefit from tighter bounds with increased $n$, but continuous classifiers offer statistical efficiency by retaining more information from $F$. Practical parameter choices (e.g., $n$, $\sigma$) balance robustness, accuracy, and computational costs based on application requirements.

\subsection{Practical Calculation of Confidence Interval}\label{subsec:practical-calculation-of-confidence-interval}

To estimate certified radii, the core task involves calculating the prediction margin $M(\Hat{F},x)$ or $M(\Phi^{-1}\circ\Hat{F},x)$ using exact confidence intervals with Bonferroni correction, as outlined in Algorithm~\ref{alg:bonferroni}. Exact intervals guarantee robustness under high confidence, unlike approximate intervals which risk overestimation in extreme probabilities or small samples.

\begin{algorithm}[h]
    \DontPrintSemicolon
    \KwIn{Counts or grid matrix $X$ and \texttt{getConfidenceBound()} routine with confidence $1-\alpha$}
    \KwOut{Margin $M(\Hat{F},x)$ or $M(\Phi^{-1}\circ\Hat{F},x)$ at confidence $1-\alpha$}

    Compute $\underline{\Hat{F}(x)}_{y}$ for the true class using \texttt{getConfidenceBound()} at $\frac{\alpha}{m}$\;
    \For{all other classes $j \neq y$}{
        Compute $\overline{\Hat{F}(x)}_{j}$ using \texttt{getConfidenceBound()} at $\frac{\alpha}{m}$\;
    }
    Calculate $M(\Hat{F},x)$ or $M(\Phi^{-1}\circ\Hat{F},x)$\;
    \KwRet{$M(\Hat{F},x)$ or $M(\Phi^{-1}\circ\Hat{F},x)$}
    \caption{Bonferroni approach for margin estimation}\label{alg:bonferroni}
\end{algorithm}

This conservative method ensures robustness guarantees by minimizing the confidence of the primary prediction while maximizing potential alternatives, providing a worst-case robustness scenario. In the discrete case, this focus is on the second-highest count due to the absence of variance within counts, ensuring a conservative robustness estimate.

Formally, for the second-highest class $j$:
\begin{align}
    M(\hat{F},x) &= \hat{F}(x)_{y} - \hat{F}(x)_j, \notag \\
    M(\Phi^{-1} \circ \hat{F}, x) &= \Phi^{-1} \big( \hat{F}(x)_{y} \big) - \Phi^{-1} \big( \hat{F}(x)_j \big).
    \label{eq:metric}
\end{align}

Using the Clopper-Pearson interval, let $n_y$ and $n_j$ represent the counts for the predicted and runner-up classes:
\begin{gather*}
    \underline{p_y} = \text{BetaInv}(\alpha/2, n_y, n - n_y + 1), \quad
    \overline{p_j} = \text{BetaInv}(1-\alpha/2, n_j + 1, n - n_j)
\end{gather*}
\begin{equation}
    \Hat{M}(\Hat{F},x)= \underline{p_y} - \overline{p_j}, \quad \Hat{M}(\Phi^{-1}\circ\Hat{F},x) = \Phi^{-1}(\underline{p_y}) - \Phi^{-1}(\overline{p_j})\label{eq:clopper-pearson}
\end{equation}

In the continuous case, confidence intervals are derived by inverting concentration inequalities. Variance-adaptive methods, like Bernstein's inequality, yield tighter bounds in low-variance scenarios, whereas empirical Bernstein inequality estimates variance from samples, avoiding overly wide intervals seen with variance-agnostic methods like Hoeffding's.

These methods provide strong robustness guarantees by underestimating the true margin, ensuring conservative robustness certificates even under probabilistic settings.

\begin{proposition}[Empirical Bernstein Inequality,~\cite{maurer2009empirical}]
    \label{prop:empirical-bernstein-inequality}
    Let $X_1, \dots, X_n$ be independent random variables in $[a,b]$, and $\bar{X}_n = \frac{1}{n}\sum_{i=1}^n X_i$ their empirical mean. The empirical variance is defined as:
    \[
        V_n = \frac{1}{n-1} \sum_{i=1}^n (X_i - \bar{X}_n)^2.
    \]
    For any $\delta \in (0,1)$ and $n \geq 2$, with probability at least $1-\delta$:
    \[
        \mathbb{E}[X] \leq \bar{X}_n + \sqrt{\frac{2V_n \ln(1/\delta)}{n}} + \frac{7(b-a)\ln(1/\delta)}{3(n-1)}.
    \]
\end{proposition}

\section{Certified Radius Estimation in the Discrete Case}\label{sec:discrete}

\subsection{Definitions and Notations}\label{subsec:definitions-and-notations}
Let $X = (X_1, \dots, X_m)$ follow a multinomial distribution with parameters $n$ and $p = (p_1, \dots, p_m)$, where $n$ is fixed and $p_k = \mathbb{P}(f(x + \epsilon) = k)$. The goal is to estimate a lower confidence bound for $\theta \coloneqq g(p)$, assuming $p \in \chi^{m-1} \subset \Delta^{m-1}$. 

The maximum likelihood estimator (MLE) of $\theta$ is $\hat{\theta}=g(\hat{p})$, where $\hat{p} = \frac{X}{n}$. Define $\Theta \coloneqq g(\chi^{m-1})$ and $\hat{\Theta} \coloneqq g(\chi^{m-1}_n)$. For observed $\tilde{\theta} \in \hat{\Theta}$, denote the cumulative distribution function (CDF) of $\hat{\theta}$ by $\Pi(\cdot|p)$:
\begin{equation}
    \Pi(L) \coloneqq 1 - \inf_{\substack{p \in \Delta^{m-1} \\ g(p) \leq L}} \Pi(\tilde{\theta}|p).\label{eq:Pi}
\end{equation}

Simplifying for $m \geq 3$, reduce the multinomial parameter to $q \in \Delta^2$ with $q_1 = p_1$, $q_2 = p_2$, and $q_3 = 1-p_1-p_2$. If $g$ depends only on $p_1$ and $p_2$, then:
\[
    \Pi(\tilde{\theta}|p) = \Pi(\tilde{\theta}|q) = \sum_{\substack{x \in \Omega^{2}_n \\ g\left(\frac{x}{n}\right) \leq \tilde{\theta}}} \binom{n}{x}q^x.
\]
The lower confidence bound on $\theta$ at level $1-\alpha$ is:
\begin{equation}
    \underline{\hat{\theta}} = \inf\{L \in \Theta : \Pi(L) = \alpha\}.
\label{eq:lower-confidence-bound}
\end{equation}

\subsection{First Radius Estimation}\label{subsec:first-radius-estimation}
For the first radius, $\chi^{m-1} = \Delta^{m-1}$, so $\Theta = [-1, 1]$ and $\hat{\Theta} = \{-1, -1 + \frac{1}{n}, \dots, 1\}$. Assuming the top two probabilities correspond to classes 1 and 2, the CDF simplifies to:
\begin{align}
    \Pi(\tilde{\theta} \mid p) &= \mathbb{P}(X_1 - X_2 \leq k \mid q) \\
    &= \sum_{x_2=0}^n \sum_{x_1=0}^{\min(k + x_2, n)} \binom{n}{x} q^x.\label{eq:pi-theta}
\end{align}

The lower confidence bound $\underline{\hat{\theta}}$ is computed by solving optimization problems in Eqs.~\eqref{eq:Pi} and~\eqref{eq:lower-confidence-bound}. The high-level algorithm is:

\begin{algorithm}[h]
    \DontPrintSemicolon
    \KwIn{Counts $X$, threshold $\epsilon>0$, routine \texttt{SolveSignomial(L)}}
    \KwOut{Lower confidence bound $\underline{\hat{\theta}}$}
    Compute $\tilde{\theta} \leftarrow \frac{X_1 - X_2}{n}$\;
    Set bounds: $\text{left} \leftarrow 0$, $\text{right} \leftarrow \tilde{\theta}$\;
    \While{$\text{right} - \text{left} > \epsilon$}{
        $L \leftarrow \frac{\text{right} + \text{left}}{2}$\;
        $\text{Solution} \leftarrow$ \texttt{SolveSignomial(L)}\;
        \If{$\text{Solution} < 1-\alpha$}{
            $\text{right} \leftarrow L$\;
        }\Else{
            $\text{left} \leftarrow L$\;
        }
    }
    \KwRet{$\text{left}$}
    \caption{First Radius Estimation in the Discrete Case}
    \label{alg:first-margin-estimation}
\end{algorithm}

For efficiency, approximate solutions can be used in early iterations, as justified by Lemma. A faster version of the algorithm is:

\begin{algorithm}[h]
    \DontPrintSemicolon
    \KwIn{Counts $X$, threshold $\epsilon>0$, routines \texttt{SolveSignomial(L)} and \texttt{FastSolveSignomial(L)}}
    \KwOut{Lower confidence bound $\underline{\hat{\theta}}$}
    Compute $\tilde{\theta} \leftarrow \frac{X_1 - X_2}{n}$\;
    Set bounds: $\text{left} \leftarrow 0$, $\text{right} \leftarrow \tilde{\theta}$, $\text{close} \leftarrow \text{False}$\;
    \While{$\text{right} - \text{left} > \epsilon$}{
        $L \leftarrow \frac{\text{right} + \text{left}}{2}$\;
        \If{$\text{close}$}{
            $\text{Solution} \leftarrow$ \texttt{SolveSignomial(L)}\;
        }\Else{
            $\text{Solution} \leftarrow$ \texttt{FastSolveSignomial(L)}\;
        }
        \If{$\text{Solution} < 1-\alpha$}{
            $\text{right} \leftarrow L$\;
        }\Else{
            $\text{close} \leftarrow \text{True}$\;
            $\text{left} \leftarrow L$\;
        }
    }
    \KwRet{$\text{left}$}
    \caption{Fast Radius Estimation in the Discrete Case}
    \label{alg:first-margin-estimation-fast}
\end{algorithm}
\subsection{Second Radius Estimation}\label{subsec:second-radius-estimation}
We solve the following optimization problem:
\begin{equation}
    \inf_{\substack{q \in \chi^{2} \\
    \Phi^{-1}(q_1) - \Phi^{-1}(q_2) \leq L}} \Pi(\Tilde{\theta} | q).\label{eq:non-signomial-optimization}
\end{equation}

Since $\Phi^{-1}$ is not a posynomial or signomial function, the problem \eqref{eq:non-signomial-optimization} cannot be solved directly. Using Taylor series approximation $\Phi^{-1}_M$, we reformulate the problem as:
\begin{align}
    \inf_{\substack{q \in \chi^2 \\
    \Phi^{-1}_M(q_1) - \Phi^{-1}_M(q_2) \leq L}} \Pi(\tilde{\theta} | q),
\end{align}
where $\Pi(\tilde{\theta} | q)$ is redefined as:
\begin{align}
    \sum_{\substack{x \in \Omega^2_n \\
    \Phi^{-1}_M\left(\frac{x_1}{n}\right) - \Phi^{-1}_M\left(\frac{x_2}{n}\right) \leq \tilde{\theta}}} \binom{n}{x} q^x. 
\end{align}

Algorithm~\ref{alg:second-margin-estimation} summarizes the procedure:
\begin{algorithm}[h]
    \DontPrintSemicolon
    \KwIn{Counts $X$, threshold $\epsilon > 0$, routines \texttt{SolveSignomial(L)} and \texttt{FastSolveSignomial(L)}}
    \KwOut{Lower confidence bound $\underline{\Hat{\theta}}$ at confidence $1-\alpha$}
    Compute $\underline{p_1}$ using the Clopper-Pearson interval at level $\frac{\alpha}{2}$.
    \If{$\underline{p_1} > \frac{1}{2}$}{
        Compute $\Tilde{\theta} \gets \Phi^{-1}\left(\frac{X_1}{n}\right) - \Phi^{-1}\left(\frac{X_2}{n}\right)$.
        Solve the problem \eqref{eq:approximated-signomial-optimization}.
    }
    \Else{
        Use the Bonferroni algorithm for standard confidence bounds.
    }
    \caption{Second Margin Estimation in the Discrete Case}\label{alg:second-margin-estimation}
\end{algorithm}

\section{Certified Radius Estimation in the Continuous Case}\label{sec:continuous}
\subsection{Estimating by Betting (Background)}\label{subsec:estimating-by-betting}
\cite{smith2022estimating} proposed confidence sequences (CS) for bounded random variables, allowing continuous monitoring with tighter bounds than empirical Bernstein's inequality. The CS framework uses:
\begin{proposition}[\cite{smith2022estimating}]
\label{prop:confidence-sequence}
For i.i.d. random variables $(X_t)_{t=1}^\infty$, a $(1-\alpha)$-CS for $\mu$ is:
\[
C_t^{\text{PrPI-EB}} \coloneqq \left( \frac{\sum_{i=1}^t \lambda_i X_i}{\sum_{i=1}^t \lambda_i} \pm \sqrt{\frac{2\log(2/\alpha) + \sum_{i=1}^t v_i \psi_e(\lambda_i)}{\sum_{i=1}^t \lambda_i}} \right),
\]
where $\lambda_t^{\text{PrPI-EB}}$, $\hat{\sigma}_t^2$, $\hat{\mu}_t$, $\psi_e(\lambda)$, and $v_i$ are defined in \eqref{eq:definitions}.
\end{proposition}

\subsection{First Radius Estimation}\label{subsec:first-radius-estimation-continuous}
To certify the smoothed classifier, define $Z \coloneqq X^{1} - \max_{j \neq 1} X^{j}$, where each $Z_i = X^{1}_i - \max_{j \neq 1} X^{j}_i$. The mean $\Bar{Z}$ is estimated conservatively using confidence intervals, applying Bonferroni correction or confidence sequences.

\subsection{Second Radius Estimation}\label{subsec:second-radius-estimation-continuous}
For the second margin, define $Z \coloneqq \Phi^{-1}(X^1) - \max_{j \neq 1}\Phi^{-1}(X^j)$. To handle unbounded $\Phi^{-1}$, we use Taylor approximation $\Phi^{-1}_M$, ensuring boundedness. If $\Bar{X^1} \geq \frac{1}{2}$, the Taylor-based method is applied; otherwise, fallback to Bonferroni correction.

Lemma ensures approximation conservativeness. This process is detailed in Algorithm~\ref{alg:second-margin-estimation}.

\section{Experiments}\label{sec:experiments}

\subsection{Certified Test-set Accuracy}\label{subsec:certified-test-set-accuracy}

Certified radius measures local classifier robustness but doesn't capture the entire input space. The \textit{certified test-set accuracy} (CTA) is a global measure for smoothed classifiers. Given a classifier \( g \), test set \( S = \{(x_1, c_1), \ldots, (x_N, c_N)\} \), and radius \( r \), we define:

\[
    z_i(r) = \mathbbm{1}[g(x_i + \delta) = c_i \quad \forall \|\delta\|_2 < r]
\]

The theoretical CTA is:

\[
    \text{CertAcc}_{\text{theo}}(r) = \frac{1}{N} \sum_{i=1}^N z_i(r)
\]

For randomized smoothing classifiers, the theoretical CTA is approximated by:

\[
    \text{CertAcc}_{\text{approx}}(r) = \frac{1}{N} \sum_{i=1}^N Y_i(r)
\]

where \( Y_i(r) \) denotes whether the certified radius of the smoothed classifier exceeds \( r \). A one-sided confidence interval can be constructed for the unobserved \( \frac{1}{N} \sum_{i=1}^N z_{i}(r) \).

\subsection{Results}\label{subsec:results}

We evaluate our radius estimation methods on the CIFAR-10 dataset. The base classifier is a 110-layer residual network~\cite{cohen2019certified} with Gaussian noise data augmentation. In all figures, CTA curves are based on margins instead of radii due to their direct proportionality.

\begin{figure}[h]
    \centering
    \includegraphics[width=\linewidth]{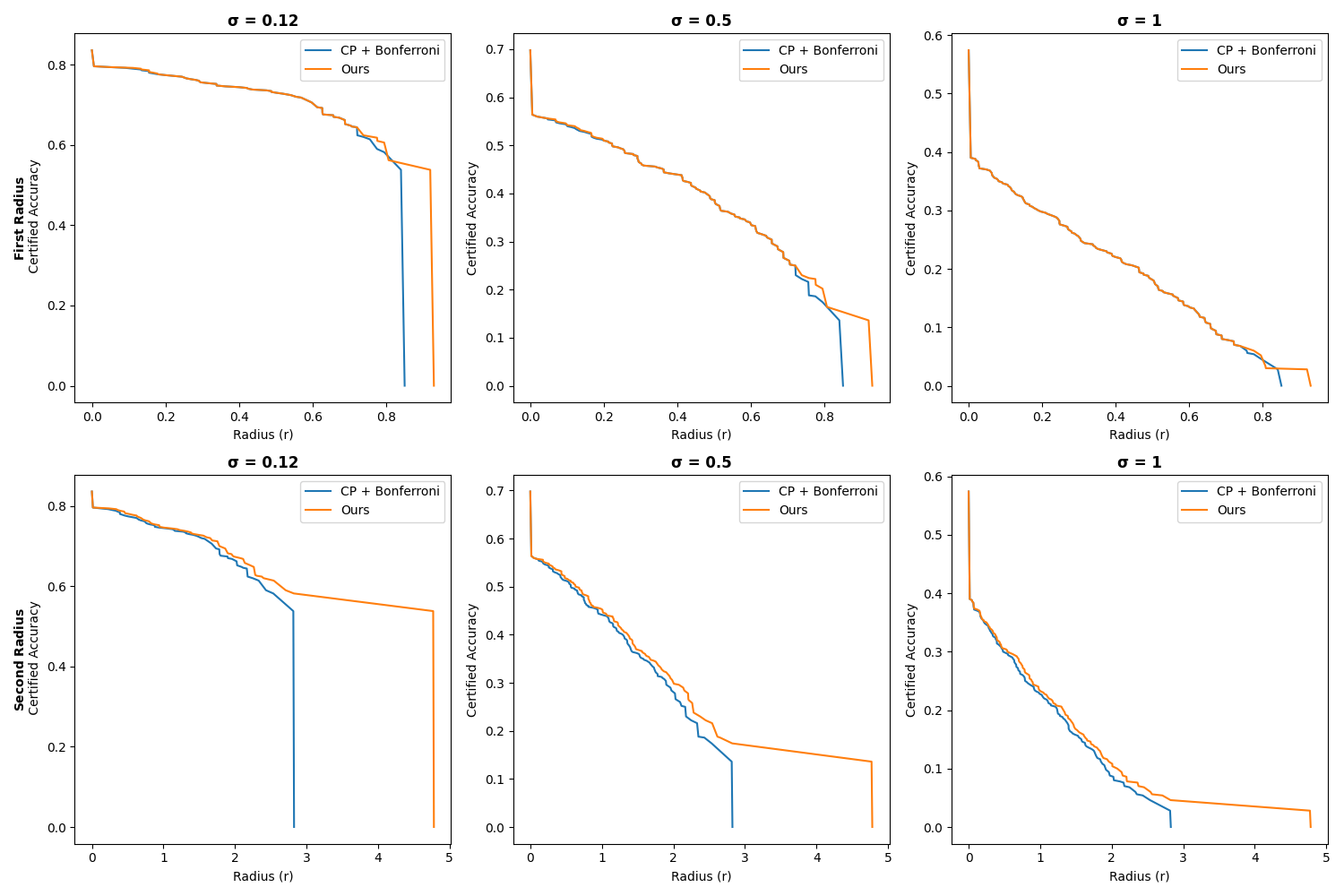}
    \caption{Certified accuracies' comparison on the CIFAR-10 dataset in the discrete case for different standard deviations (displayed on the columns) with a sample size of $100$. The legend and row conventions are the same as in Figure~\ref{fig:discrete_num}.}
    \label{fig:discrete_sigma}
\end{figure}

Figure~\ref{fig:discrete_sigma} compares certified accuracies on CIFAR-10 for various standard deviations. The CTA decreases as the standard deviation increases, as higher noise reduces the base accuracy, making classification harder.

\begin{figure}[h]
    \centering
    \includegraphics[width=\linewidth]{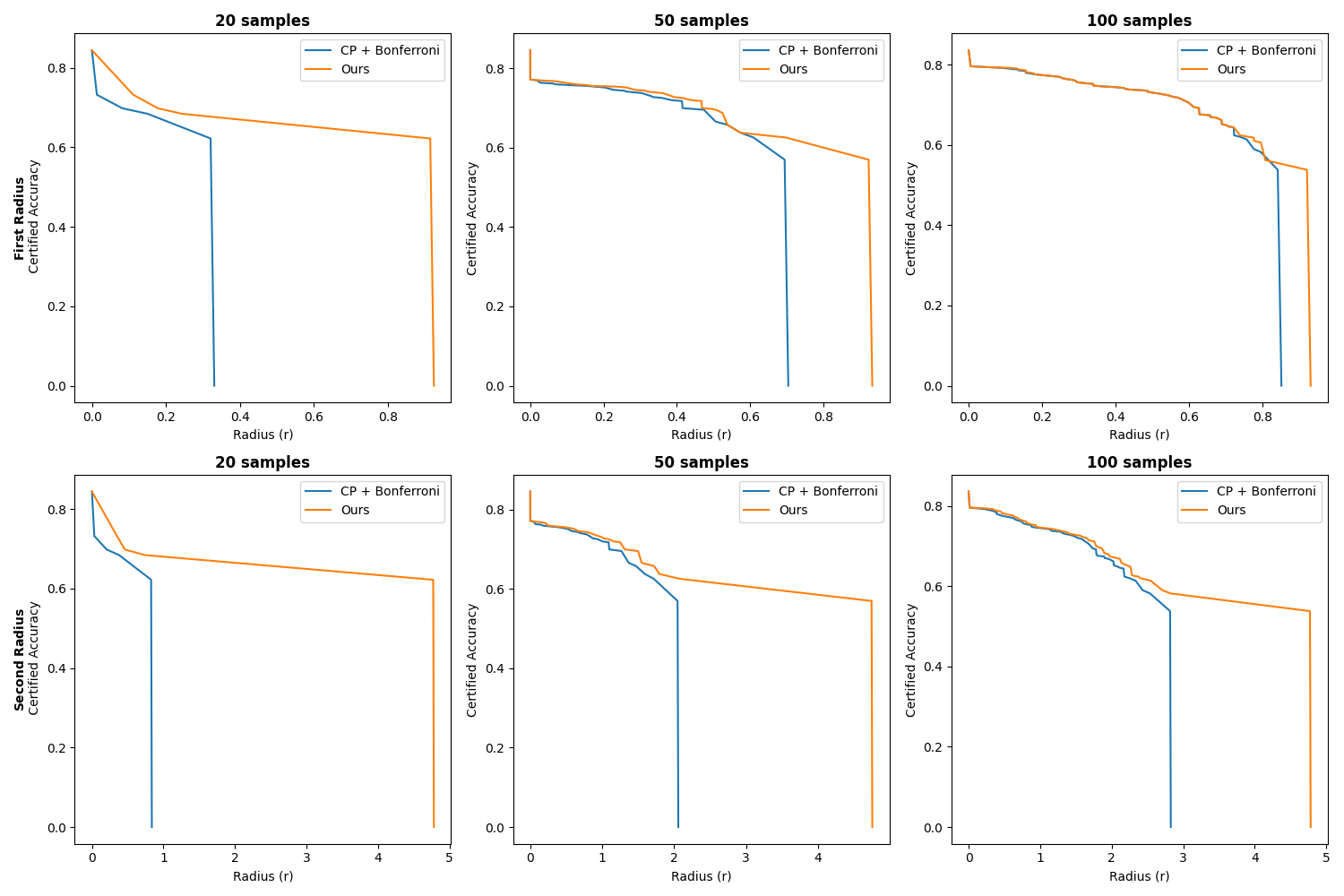}
    \caption{Certified accuracies' comparison on the CIFAR-10 dataset in the discrete case for different numbers of samples (displayed on the columns) with $\sigma = 0.12$. CP + Bonferroni means Clopper-Pearson interval with Bonferroni correction, and Ours means our new approach in section~\ref{sec:discrete}. The first row compares the certified accuracies using the first margin and the second row compares the certified accuracies using the second margin.}
    \label{fig:discrete_num}
\end{figure}

Figure~\ref{fig:discrete_num} and Table~\ref{tab:simplified-certified-accuracy} show CTA results for different sample sizes and the impact of sample size and standard deviation on the CTA curve. The standard Bonferroni approach is conservative for small sample sizes, but as the sample size grows, the two curves converge. Differences are more pronounced at larger radii.

\begin{table*}[h]
    \centering
    \caption{Certified accuracy using the second margin on the CIFAR-10 dataset in the discrete case for different values of radius $r$ with a sample size of $100$ and a standard deviation $\sigma = 0.12$.}
    \label{tab:simplified-certified-accuracy}
    \renewcommand{\arraystretch}{1.2}
    \begin{tabular}{l*{9}{c}}
        \toprule
        Method & \multicolumn{9}{c}{Radius ($r$)} \\
        \cmidrule(l){2-10}
        & 0.5 & 1.0 & 1.5 & 2.0 & 2.5 & 3.0 & 3.5 & 4.0 & 4.5 \\
        \midrule
        CP + Bonferroni & 0.774 & 0.744 & 0.720 & 0.662 & 0.582 & 0.000 & 0.000 & 0.000 & 0.000 \\
        Ours              & 0.780 & 0.746 & 0.726 & 0.670 & 0.614 & 0.538 & 0.538 & 0.538 & 0.538 \\
        Gain (\%) & 0.78\% & 0.27\% & 0.83\% & 1.21\% & 5.50\% & $\infty$ & $\infty$ & $\infty$ & $\infty$ \\
        \bottomrule
    \end{tabular}
\end{table*}

In the continuous case, in addition to the sample size and the standard deviation, another hyperparameter to consider is the \textit{temperature}.
The simplex map $s$ used in the experiments is the tempered softmax function which is a generalization of the standard softmax function, introducing a temperature parameter to control the smoothness of the output distribution.
Given a vector $\mathbf{x} = (x_1, \ldots, x_m)$ and a temperature parameter $T > 0$, the tempered softmax function $\sigma_T: \mathbb{R}^m \to \mathbb{R}^m$ is defined as:
\[
    \sigma_T(\mathbf{x})_i = \frac{\exp(x_i/T)}{\sum_{j=1}^n \exp(x_j/T)}
\]
for $i = 1, \ldots, m$.

As $T \to 0^+$, the tempered softmax approaches a hard maximum (one-hot vector).
As $T \to \infty$, the tempered softmax approaches a uniform distribution.
When $T = 1$, it reduces to the standard softmax function.

\begin{figure}[h]
    \centering
    \includegraphics[width=\linewidth]{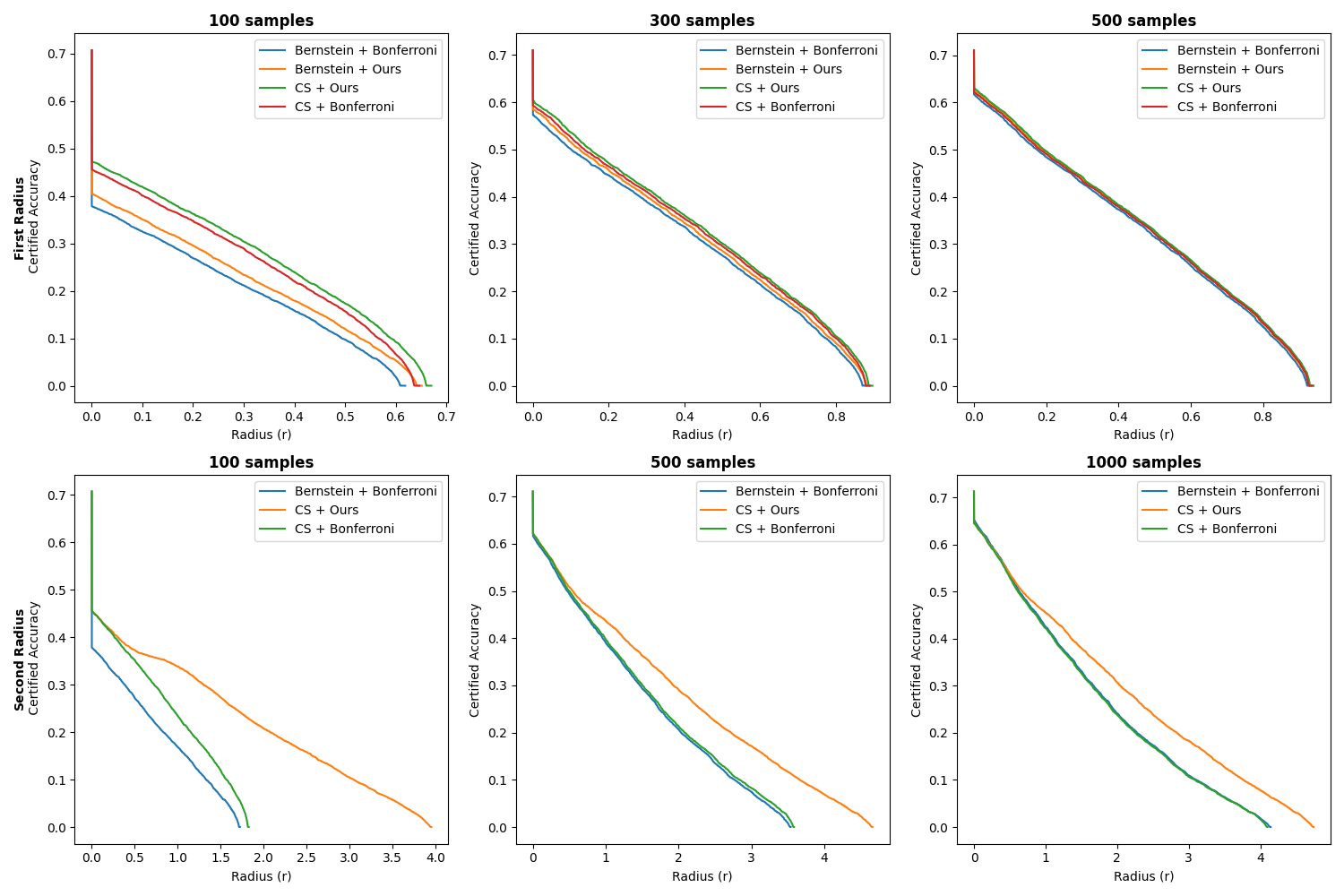}
    \caption{Certified accuracies' comparison on the CIFAR-10 dataset in the continuous case for different sample sizes (displayed on the columns) with $\sigma = 0.5$ and a temperature equal to $1$. CS/Bernstein + Bonferroni stands for the Bonferroni approach with either the empirical Bernstein interval (Proposition~\ref{prop:empirical-bernstein-inequality}) or the confidence sequence (Proposition~\ref{prop:confidence-sequence}), and CS/Bernstein + Ours stands for the new approach in section~\ref{sec:continuous}, where the interval used is either the empirical Bernstein interval or the confidence sequence as before. The first row compares the certified accuracies using the first margin and the second row compares the certified accuracies using the second margin.}
    \label{fig:cont_num}
\end{figure}

\begin{figure}[h]
    \centering
    \includegraphics[width=\linewidth]{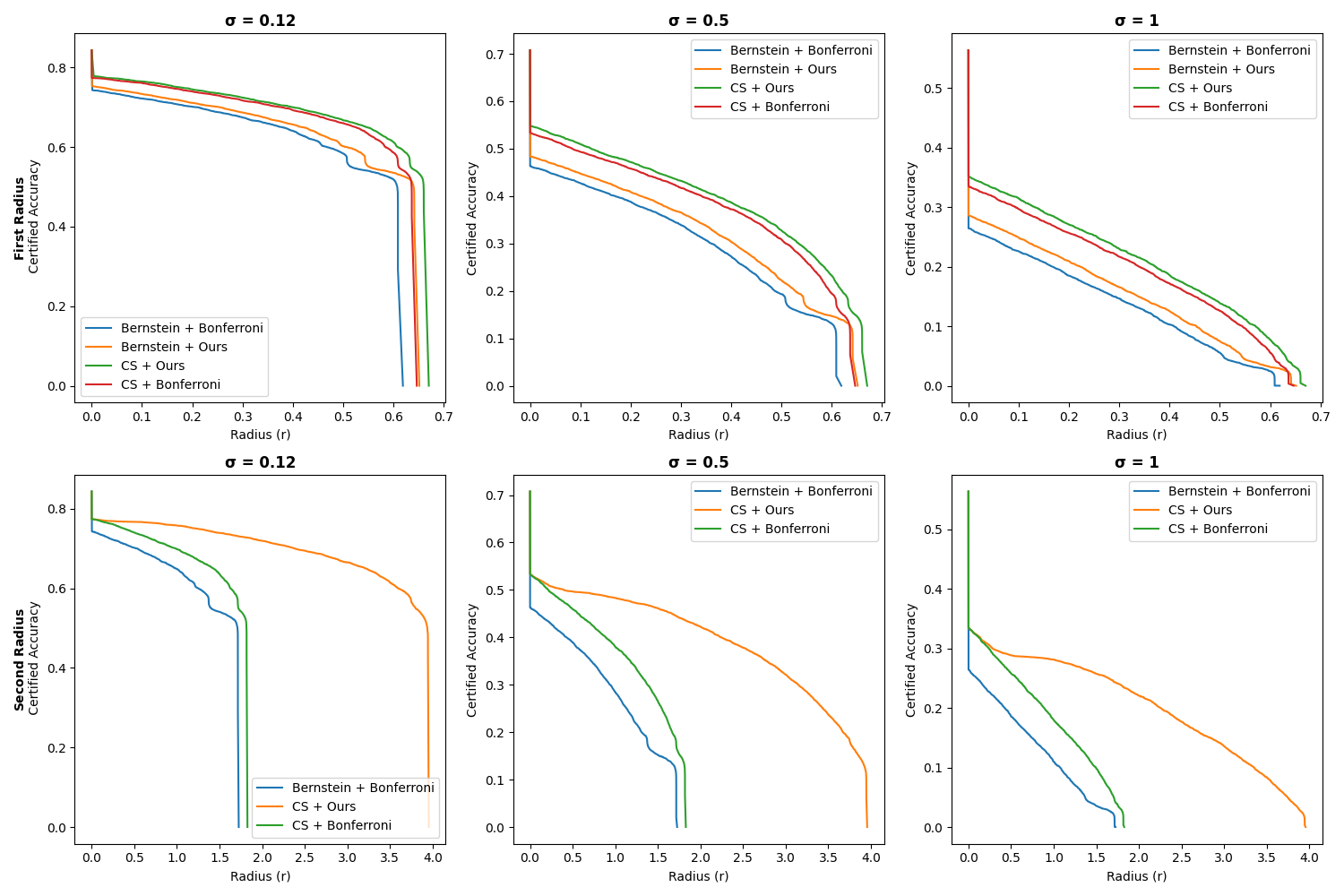}
    \caption{Certified accuracies' comparison on the CIFAR-10 dataset in the continuous case for different standard deviations (displayed on the columns) with a sample size of $100$ and a temperature equal to $0.1$. The legend and row conventions are the same as in Figure~\ref{fig:cont_num}.}
    \label{fig:cont_sigma}
\end{figure}

Figures~\ref{fig:cont_num} \&~\ref{fig:cont_sigma} and tables~\ref{tab:certified-accuracy} \&~\ref{tab:certified-accuracy-2} show the effect of increasing the number of samples $n$ and increasing the standard deviation $\sigma$ respectively.
The effects of these hyperparameters are similar in both discrete and continuous case.

\begin{table*}[h]
    \centering
    \caption{Certified accuracy using the first margin on CIFAR-10 in the continuous case for different values of radius $r$ and sample sizes with $\sigma = 0.5$ and a temperature of $1$.}
    \label{tab:certified-accuracy}
    \renewcommand{\arraystretch}{1.2}
    \begin{tabular}{@{}ll*{9}{c}@{}}
        \toprule
        \multirow{2}{*}{Samples} & \multirow{2}{*}{Method} & \multicolumn{9}{c}{Radius ($r$)} \\
        \cmidrule(l){3-11}
        & & 0.1 & 0.2 & 0.3 & 0.4 & 0.5 & 0.6 & 0.7 & 0.8 & 0.9 \\
        \midrule
        \multirow{3}{*}{100}
        & CS + Bonferroni & 0.400 & 0.347 & 0.289 & 0.220 & 0.157 & 0.066 & 0.000 & 0.000 & 0.000 \\
        & CS + Ours            & 0.419 & 0.362 & 0.303 & 0.239 & 0.173 & 0.092 & 0.000 & 0.000 & 0.000 \\
        & Comparison (\%) & 4.69\% & 4.28\% & 5.04\% & 8.70\% & 10.65\% & 39.16\% & N/A & N/A & N/A \\
        \midrule
        \multirow{3}{*}{300}
        & CS + Bonferroni & 0.526 & 0.464 & 0.411 & 0.354 & 0.296 & 0.233 & 0.173 & 0.101 & 0.000 \\
        & CS + Ours             & 0.536 & 0.471 & 0.416 & 0.361 & 0.301 & 0.239 & 0.179 & 0.105 & 0.000 \\
        & Comparison (\%) & 1.93\% & 1.55\% & 1.39\% & 1.89\% & 1.94\% & 2.52\% & 3.49\% & 4.06\% & N/A \\
        \midrule
        \multirow{3}{*}{500}
        & CS + Bonferroni & 0.560 & 0.490 & 0.434 & 0.379 & 0.322 & 0.261 & 0.199 & 0.133 & 0.050 \\
        & CS + Ours             & 0.568 & 0.497 & 0.442 & 0.383 & 0.329 & 0.266 & 0.202 & 0.136 & 0.054 \\
        & Comparison (\%) & 1.43\% & 1.33\% & 2.00\% & 0.99\% & 2.08\% & 2.01\% & 1.89\% & 2.33\% & 9.21\% \\
        \bottomrule
    \end{tabular}
\end{table*}
\begin{table*}[h]
    \centering
    \caption{Certified accuracy using the second margin on CIFAR-10 in the continuous case for different values of radius $r$ and sample sizes with $\sigma = 0.5$ and a temperature of $1$.}
    \label{tab:certified-accuracy-2}
    \renewcommand{\arraystretch}{1.2}
    \begin{tabular}{@{}ll*{9}{c}@{}}
        \toprule
        \multirow{2}{*}{Samples} & \multirow{2}{*}{Method} & \multicolumn{9}{c}{Radius ($r$)} \\
        \cmidrule(l){3-11}
        & & 0.1 & 0.2 & 0.3 & 0.4 & 0.5 & 0.6 & 0.7 & 0.8 & 0.9 \\
        \midrule
        \multirow{3}{*}{100}
        & CS + Bonferroni & 0.436 & 0.415 & 0.393 & 0.369 & 0.352 & 0.328 & 0.305 & 0.284 & 0.259 \\
        & CS + Ours             & 0.437 & 0.418 & 0.401 & 0.383 & 0.373 & 0.364 & 0.359 & 0.354 & 0.347 \\
        & Comparison (\%) & 0.19\% & 0.83\% & 1.91\% & 3.72\% & 5.95\% & 11.17\% & 17.68\% & 24.67\% & 34.03\% \\
        \midrule
        \multirow{3}{*}{300}
        & CS + Bonferroni & 0.572 & 0.545 & 0.518 & 0.493 & 0.471 & 0.451 & 0.431 & 0.413 & 0.391 \\
        & CS + Ours             & 0.572 & 0.547 & 0.523 & 0.500 & 0.481 & 0.465 & 0.451 & 0.439 & 0.427 \\
        & Comparison (\%) & 0.00\% & 0.42\% & 0.92\% & 1.46\% & 2.15\% & 2.97\% & 4.59\% & 6.25\% & 9.24\% \\
        \midrule
        \multirow{3}{*}{500}
        & CS + Bonferroni & 0.600 & 0.578 & 0.553 & 0.525 & 0.500 & 0.478 & 0.457 & 0.437 & 0.419 \\
        & CS + Ours             & 0.600 & 0.578 & 0.556 & 0.530 & 0.509 & 0.489 & 0.473 & 0.461 & 0.448 \\
        & Comparison (\%) & 0.08\% & 0.08\% & 0.53\% & 1.03\% & 1.77\% & 2.29\% & 3.50\% & 5.62\% & 6.98\% \\
        \bottomrule
    \end{tabular}
\end{table*}

\begin{figure}[h]
    \centering
    \includegraphics[width=\linewidth]{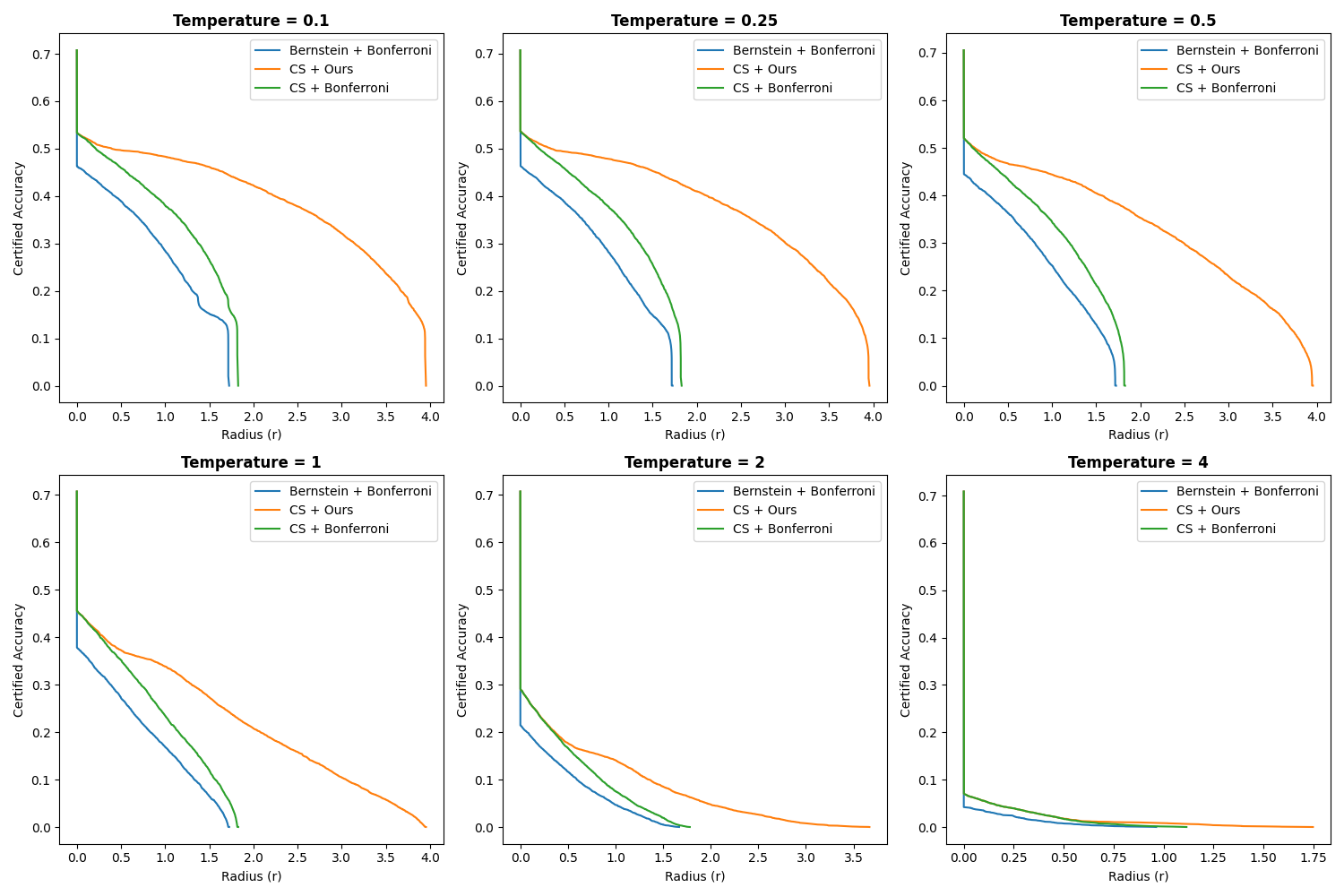}
    \caption{Certified accuracies' comparison on the CIFAR-10 dataset in the continuous case for different temperatures (displayed on the columns) with a sample size of $100$ and $\sigma = 0.5$. The legend and row conventions are the same as in Figure~\ref{fig:cont_num}.}
    \label{fig:cont_temp}
\end{figure}

It is clear from Figure~\ref{fig:cont_temp} that increasing the temperature parameter $T$ reduces the discrepancies between the Bonferroni approach and our new method.
Higher temperatures lead to softer decision boundaries.
As the temperature increases, the output probabilities of the tempered softmax function become more uniform, regardless of the input values.
This means that the function becomes less sensitive to differences in the input, causing different methods to produce more similar outputs.
Hence, as the temperature rises, the tempered softmax function becomes less responsive to changes in its inputs.
This means that larger changes in the input are required to produce the same change in output probabilities.
Consequently, the differences between various methods become less pronounced.
At lower temperatures, the tempered softmax accentuates differences between inputs.
The highest value tends to dominate, resulting in an output distribution that's closer to a one-hot vector.

\section{Conclusion and Future Work}\label{sec:conclusion-and-future-work}
In this paper, we have presented novel techniques for improving the estimation of certified radii in randomized smoothing, leading to tighter bounds on certified test-set accuracy.
Our methods have demonstrated significant improvements on both CIFAR-10 and ImageNet datasets, showcasing the potential for more efficient and accurate certification of neural network robustness against adversarial perturbations.
While our results mark a substantial step forward in the field of adversarial robustness, they also open up several promising avenues for future research:
One particularly intriguing direction is the development of more efficient tricks for estimating certified radii in discrete domains.
In the continuous case, our work has highlighted the importance of tight confidence intervals for accurate estimation of certified radii, leaving the exploration of tighter confidence sequences and the development of new theoretical frameworks that provide rigorous backing for the tightness of these improved confidence intervals for future work.
By pursuing these lines of research, we hope to further narrow the gap between empirical performance and theoretical guarantees in randomized smoothing.
We leave for future the comparison between empirical certified radii (like those based on PGD attacks) and estimated certified radii.

    \appendix

   \section{Signomial Programming}\label{sec:signomial-programming}

\subsection{Definitions}\label{subsec:definitions}
In signomial programming, the building block is a \textbf{monomial} function which is a function of the form
\[f(\mathbf{x}) = c x_1^{a_1} x_2^{a_2} \cdots x_m^{a_m}\]

where:
\begin{itemize}
    \item $c > 0$ is a positive coefficient,
    \item $\mathbf{x} = (x_1, x_2, \ldots, x_m)$ are positive variables,
    \item $a_1, a_2, \ldots, a_m$ are real exponents (not necessarily non-negative).
\end{itemize}
Building on top of this definition, we can define two types of functions.
A \textbf{posynomial} function is a sum of monomials, while a \textbf{signomial} function is a linear combination of monomials (meaning that the multiplicative coefficients can be negative).

A signomial program (SP) is an optimization problem of the form:
\[
    \begin{aligned}
        & \text{minimize}   & & f_0(\mathbf{x}) \\
        & \text{subject to} & & f_i(\mathbf{x}) \geq 0, \quad i = 1, \ldots, p \\
        &                   & & \mathbf{x} > 0
    \end{aligned}
\]
where:

\begin{itemize}
    \item $\mathbf{x} = (x_1, \ldots, x_m)$ is the vector of optimization variables,
    \item $f_0, f_1, \ldots, f_m$ are signomial functions.
\end{itemize}

Signomials programs generalize the well-known geometric programs that are much easier to solve, since they can be reduced to convex optimization problems.
However, signomial programs are generally non-convex optimization problems and can be challenging to solve globally.
Various techniques, such as successive convex approximation or branch-and-bound methods, are often employed to find solutions to SPs.

\section{Proofs}\label{sec:proofs}

\subsection{Proof of Lemma }\label{subsec:proof-of-lemma-ref{lemma:suboptimal}}
\begin{proof}
    Suppose there exists $q^0\in\Delta^{2}$ such that $q^0_1-q^0_2\leq L$ and $\Pi(\Tilde{\theta}|q^0)\leq 1-\alpha$ for some $L\in\Theta$.
    It follows that
    \[
        \inf_{\substack{q\in\Delta^{2}\\q_1-q_2\leq L}}\Pi(\Tilde{\theta}|q)\leq\Pi(\Tilde{\theta}|q^0)\leq 1-\alpha.
    \]
    In other terms,
    \[\alpha\leq\Pi(L).\]
    Since $\Pi$ is nondecreasing, by definition of $\underline{\Hat{\theta}}$,
    \[\Pi(\underline{\Hat{\theta}})\leq\alpha.\]
    Therefore, $\Pi(\underline{\Hat{\theta}})\leq\Pi(L)$, which implies $\underline{\Hat{\theta}}\leq L$.
\end{proof}

\subsection{Proof of Lemma}\label{subsec:proof-of-lemma-ref{lemma:approximation}}
\begin{proof}
    To recall, the error function, denoted as $\text{erf}(x)$, is defined as

    \[
        \text{erf}(x) = \frac{2}{\sqrt{\pi}} \int_0^x e^{-t^2} dt.
    \]

    Its domain of the error function is the interval $(-\infty, \infty)$, and its codomain is the interval $(-1, 1)$.
    The error function is an increasing and odd function.

    The inverse error function, denoted as $\text{erf}^{-1}(z)$, is the inverse of the error function.
    Its domain is the interval $(-1, 1)$, and its codomain is all real numbers.
    Due to the complexity of the error function, the inverse error function does not have a simple closed-form expression.
    However, it can be approximated using various methods.

    One such approximation for the inverse error function is given by the following series:

    \[
        \text{erf}^{-1}(x) = \sum_{k=0}^{\infty} \frac{c_k}{2k+1} \left(\frac{\sqrt{\pi}}{2}x\right)^{2k+1}
    \]

    where $c_0 = 1$ and the subsequent coefficients $c_k$ are defined recursively as:

    \[
        c_k = \sum_{m=0}^{k-1} \frac{c_m c_{k-1-m}}{(m+1)(2m+1)}.
    \]

    This series approximation converges on the entire domain of the inverse error function.
    If we denote by $\text{erf}_M$ the $M$-th order Taylor series of the error function,
    \[
        \text{erf}_M(x) \coloneqq \sum_{k=0}^M \frac{c_k}{2k+1} \left(\frac{\sqrt{\pi}}{2}x\right)^{2k+1},
    \]
    then it is clear that $\text{erf}(x) \geq\text{erf}_M(x)$ if $x\geq0$, and $\text{erf}(x) \leq\text{erf}_M(x)$ otherwise.

    The Gaussian quantile function, denoted as $\Phi^{-1}$, can be expressed in terms of the inverse error function as follows
    \[\Phi^{-1}(p) = \sqrt{2} \cdot \text{erf}^{-1}(2p - 1).\]

    The domain of $\Phi^{-1}(p)$ is $(0, 1)$, corresponding to probabilities, while its codomain is $\mathbb{R}$.
    The Taylor series approximation of $\text{erf}^{-1}(x)$ naturally leads to an approximation of the Gaussian quantile function.
    By substituting $x = 2p - 1$ into the Taylor series for $\text{erf}^{-1}(x)$ and multiplying by $\sqrt{2}$, we obtain
    \[\Phi^{-1}(p) \approx \sqrt{2} \sum_{k=0}^{M} \frac{c_k}{2k+1} \left(\frac{\sqrt{\pi}}{2}(2p-1)\right)^{2k+1}\coloneqq\Phi^{-1}_M(p)\]
    It follows that if $p\geq\frac{1}{2}$, then $\Phi^{-1}(p) \geq\Phi^{-1}_M(p)$, and if $p\leq\frac{1}{2}$, then $\Phi^{-1}(p) \leq\Phi^{-1}_M(p)$, which concludes the proof.
\end{proof}

  \section{More Experimental Results}\label{sec:more-experimental-results}
\begin{figure}[h]
    \centering
    \includegraphics[width=\linewidth]{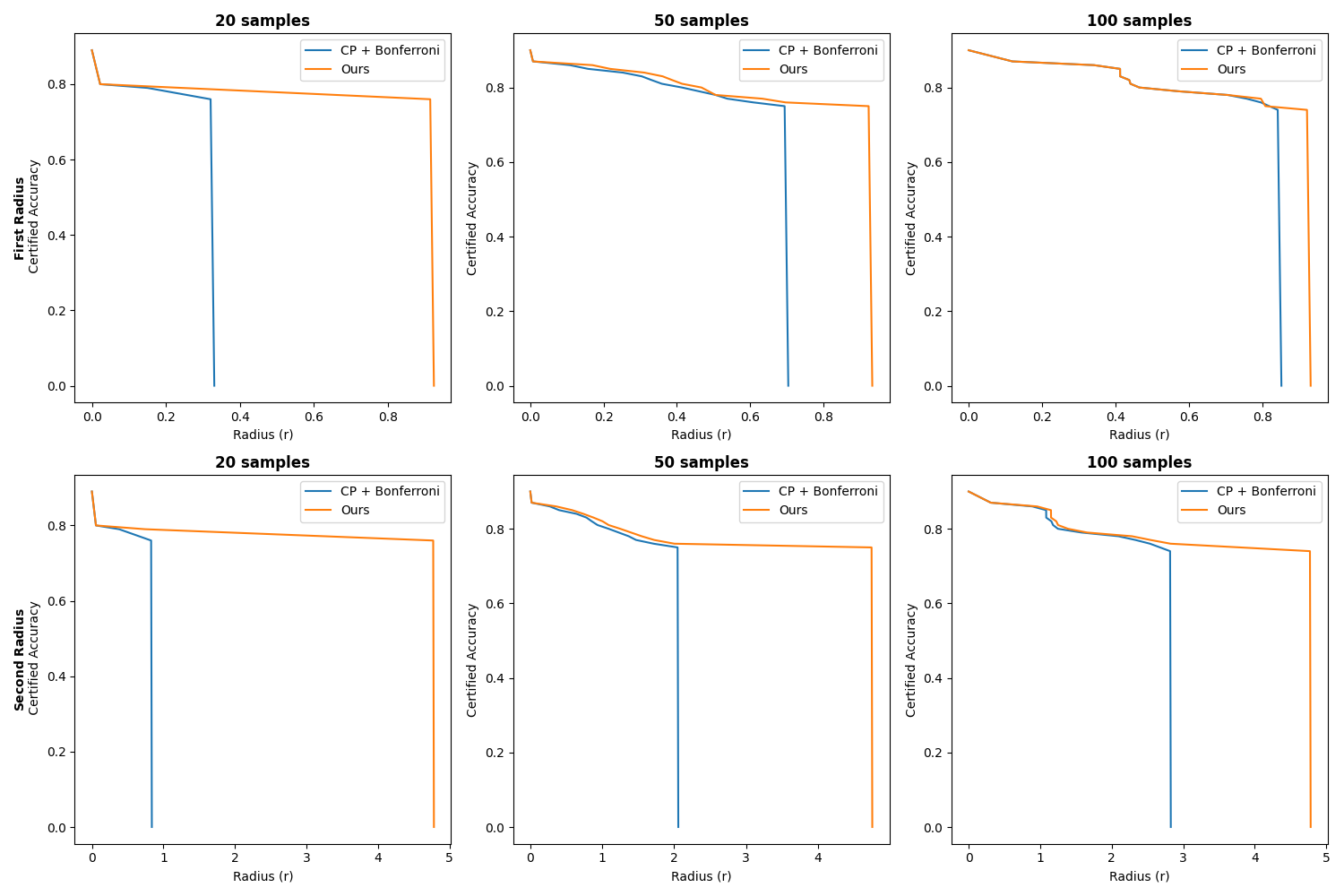}
    \caption{Certified accuracies' comparison on the ImageNet dataset in the discrete case for different numbers of samples (displayed on the columns) with $\sigma = 0.25$.}
    \label{fig:discrete_num_imagenet}
\end{figure}

\begin{figure}[h]
    \centering
    \includegraphics[width=\linewidth]{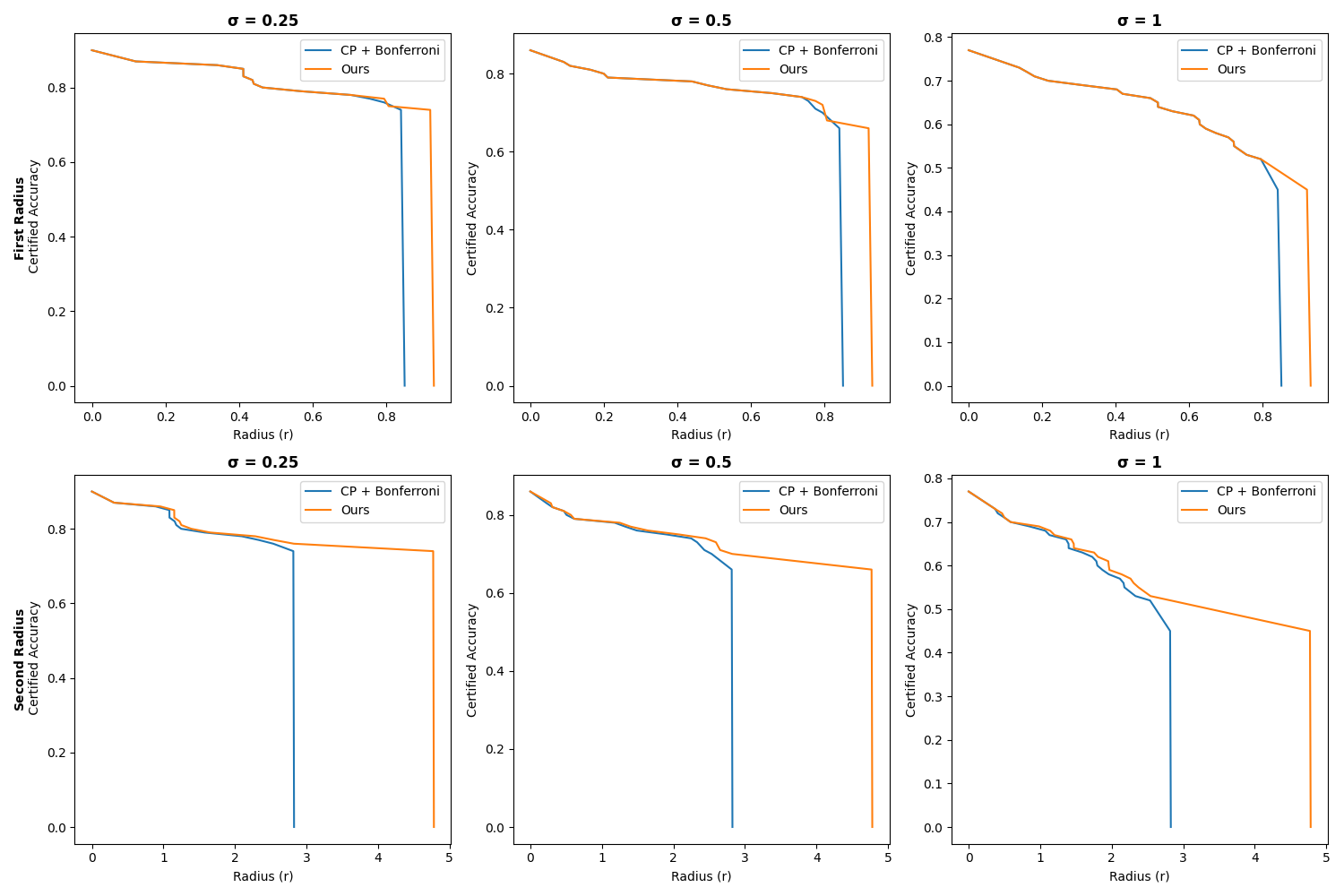}`
    \caption{Certified accuracies' comparison on the ImageNet dataset in the discrete case for different standard deviations (displayed on the columns) with a sample size of $100$.}
    \label{fig:discrete_sigma_imagenet}
\end{figure}

\begin{figure}[h]
    \centering
    \includegraphics[width=\linewidth]{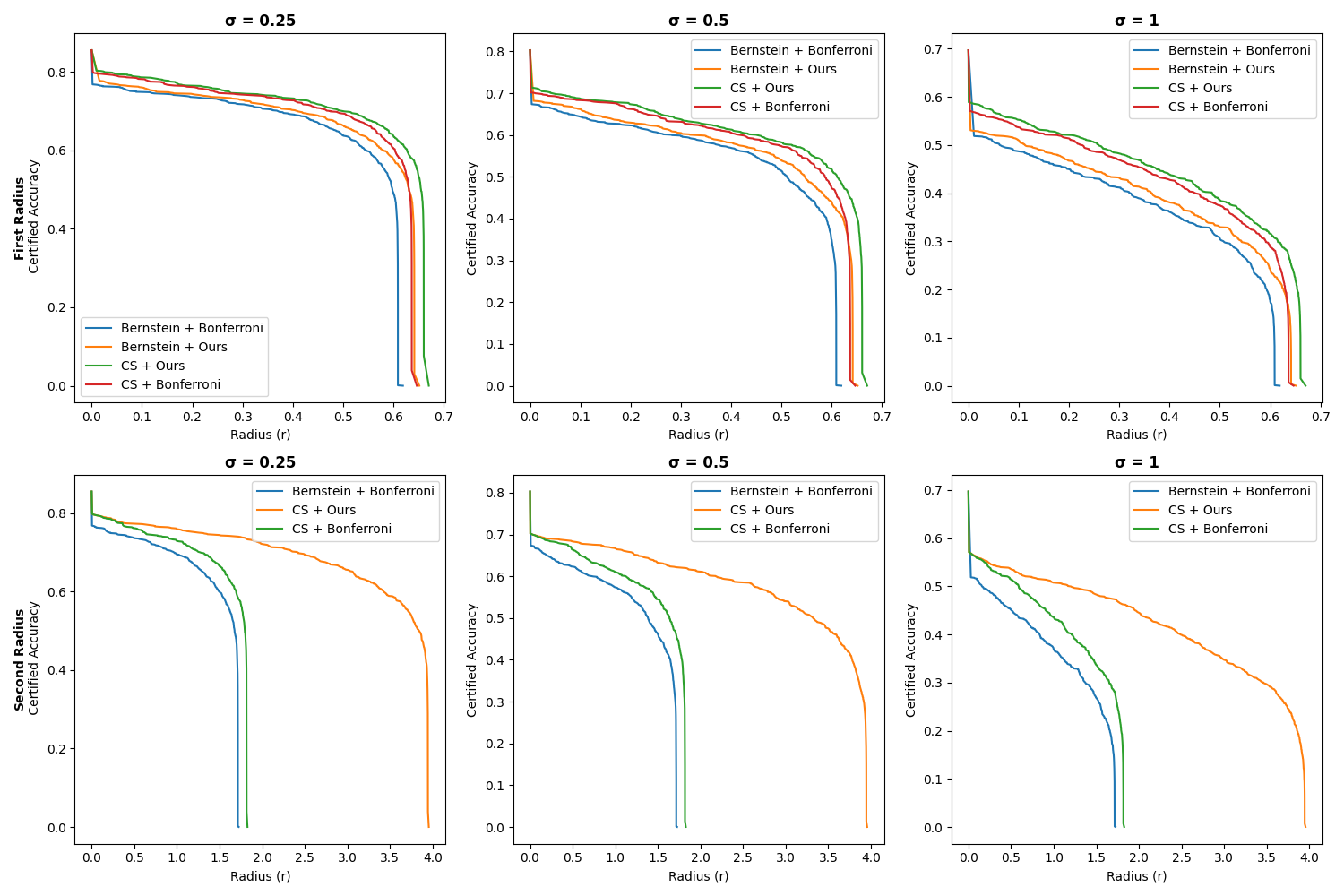}`
    \caption{Certified accuracies' comparison on the ImageNet dataset in the continuous case for different standard deviations (displayed on the columns) with a sample size of $100$ and a temperature equal to $0.5$.}
    \label{fig:cont_sigma_imagenet}
\end{figure}

\begin{figure}[h]
    \centering
    \includegraphics[width=\linewidth]{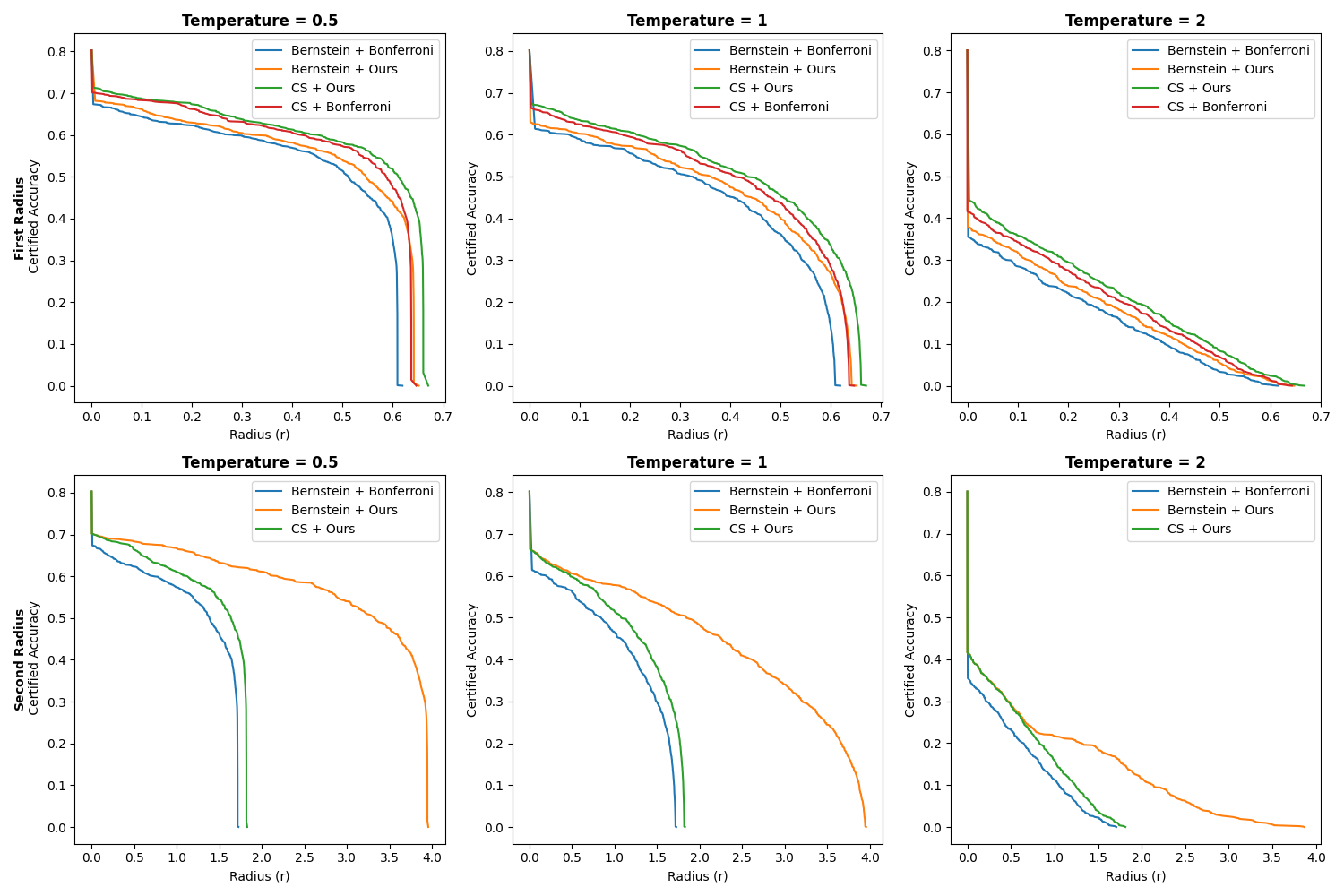}
    \caption{Certified accuracies' comparison on the CIFAR-10 dataset in the continuous case for different temperatures (displayed on the columns) with a sample size of $100$ and $\sigma = 0.5$.}
    \label{fig:cont_temp_imagenet}
\end{figure}

We conducted our experiments on the ImageNet dataset in the same way as in Section~\ref{sec:experiments}.
The only difference is that we used a 50-layer residual network instead of the 110-layer one.
The figures of the Imagenet dataset (\ref{fig:discrete_num_imagenet},~\ref{fig:discrete_sigma_imagenet},~\ref{fig:cont_sigma_imagenet}, and~\ref{fig:cont_temp_imagenet}) are consistent with our previous findings on the CIFAR-10 dataset.

    \bibliographystyle{IEEEtran}
    \bibliography{references}

\end{document}